\documentclass{article}

\usepackage[final]{corl_2019} 

\usepackage{wrapfig}
\usepackage{subcaption}
\usepackage{graphicx}
\usepackage{amsmath}
\usepackage{amsfonts}
\usepackage{mathtools}
\usepackage{array}
\usepackage{booktabs}
\usepackage{makecell}
\usepackage{multirow}
\usepackage{caption}

\usepackage{titlesec}
\titlespacing\section{0pt}{3pt plus 4pt minus 2pt}{0pt plus 2pt minus 2pt}
\titlespacing\subsection{0pt}{3pt plus 4pt minus 2pt}{0pt plus 2pt minus 2pt}
\titlespacing\subsubsection{0pt}{3pt plus 4pt minus 2pt}{0pt plus 2pt minus 2pt}

\title{RoboNet: Large-Scale Multi-Robot Learning
}

\author{
  Sudeep Dasari$^{1,4}$, Frederik Ebert$^1$, Stephen Tian$^1$, Suraj Nair$^2$, Bernadette Bucher$^3$\\
  \textbf{Karl Schmeckpeper$^3$, Siddharth Singh$^3$, Sergey Levine$^1$, Chelsea Finn$^2$} \\
  UC Berkeley$^1$, Stanford University$^2$, University of Pennsylvania$^3$,  CMU$^4$\\
}

\begin{document}
\maketitle


\begin{abstract}
Robot learning has emerged as a promising tool for taming the complexity and diversity of the real world. Methods based on high-capacity models, such as deep networks, hold the promise of providing effective generalization to a wide range of open-world environments. However, these same methods typically require large amounts of diverse training data to generalize effectively. In contrast, most robotic learning experiments are small-scale, single-domain, and single-robot. This leads to a frequent tension in robotic learning: how can we learn generalizable robotic controllers without having to collect impractically large amounts of data for each separate experiment? In this paper, we propose RoboNet, an open database for sharing robotic experience, which provides an initial pool of 15 million video frames, from 7 different robot platforms, and study how it can be used to learn generalizable models for vision-based robotic manipulation.
We combine the dataset with two different learning algorithms: visual foresight, which uses forward video prediction models, and supervised inverse models. Our experiments test the learned algorithms' ability to work across new objects, new tasks, new scenes, new camera viewpoints, new grippers, or even entirely new robots. In our final experiment, we find that by pre-training on RoboNet and fine-tuning on data from a held-out Franka or Kuka robot, we can exceed the performance of a robot-specific training approach that uses 4x-20x more data.\footnote{For videos and data, see the project webpage: \url{http://www.robonet.wiki/}}
\end{abstract}



\section{Introduction}

The key motivation for using machine learning in robotics is to build systems that can handle the diversity of open-world environments, which demand the ability to generalize to new settings and tasks. Such generalization may either be \emph{zero-shot}, without any additional data from the target domain, or very fast, using only a modest amount of target domain data.
Despite this promise, two of the most commonly raised criticisms of machine learning applied to robotics are the amount of data required per environment due to limited data-sharing, and the resulting algorithm's poor generalization to even modest environmental changes.
A number of works have tried to address this by developing simulations from which large amounts of diverse data can be collected~\cite{sadeghi2016cad2rl,andrychowicz2018learning}, or by attempting to make robot learning algorithms more data efficient~\cite{deisenroth2011pilco, deisenroth2013gaussian}. However, developing simulators entails a deeply manual process, which so far has not scaled to the breadth and complexity of open-world environments. The alternative of using less real-world data often also implies using simpler models, which are insufficient for capturing the many details present in complex real-world environments such as object geometry or appearance.

\begin{figure}[t]
    \centering
    \includegraphics[width=1\textwidth]{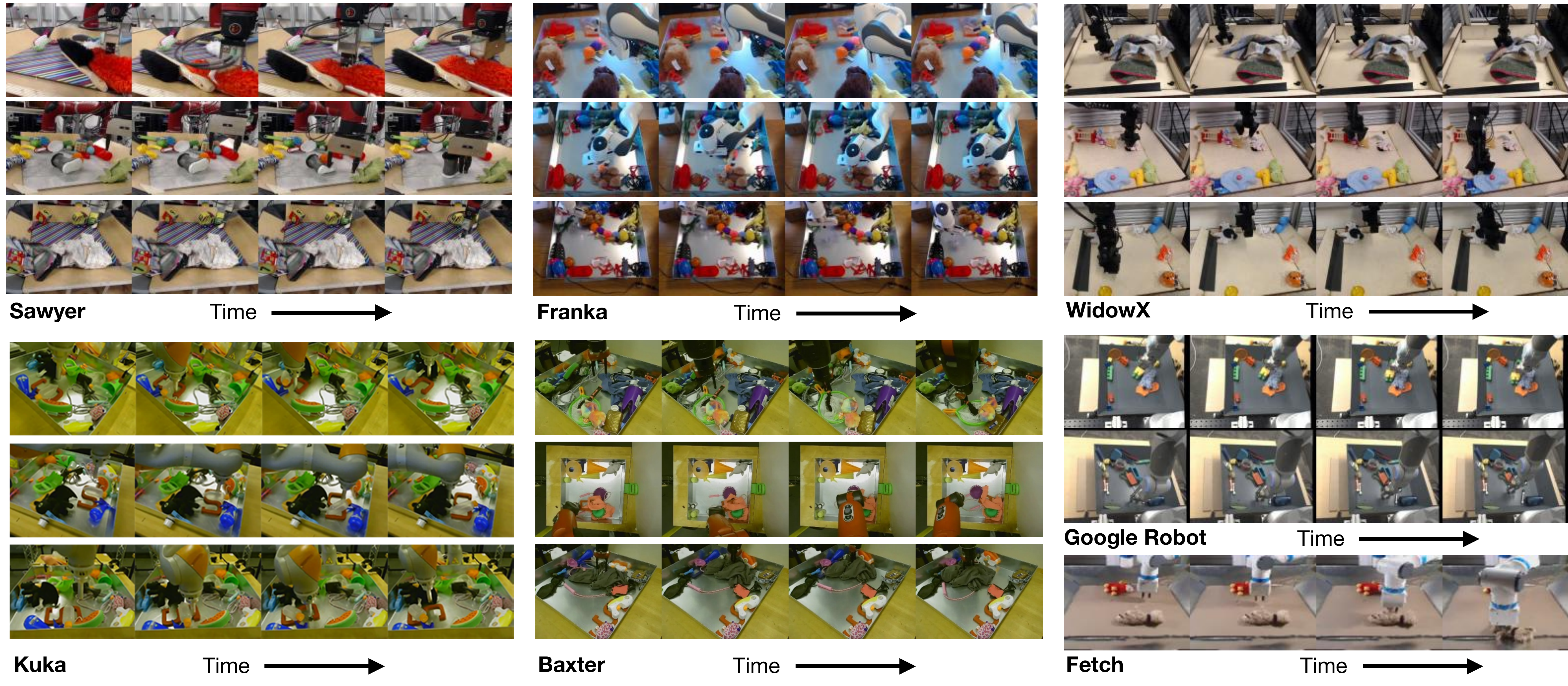}
    \caption{A glimpse of the RoboNet dataset, with example trajectories, robots, and viewpoints. We collected data with Sawyer, Franka, WidowX, Kuka, and Baxter robots, and augmented the dataset with publicly-available data from a robot from Google~\cite{finn2016unsupervised}, a Fetch~\cite{yu2019unsupervised}, and a Sawyer ~\cite{ebert2018deep}. We use RoboNet to study the viability of large-scale data-driven robot learning, as a means to attain broad generalization across robots and scenes.
    }
    \label{fig:robonet_overview}
    \vspace{-0.2in}
\end{figure}

Instead, we propose the opposite -- using dramatically larger and more varied datasets collected in the real world. Inspired by the breadth of the ImageNet dataset~\cite{deng2009imagenet}, we introduce \emph{RoboNet}, a dataset containing roughly 162,000 trajectories with video and action sequences recorded from 7 robots, interacting with hundreds of objects, with varied viewpoints and environments, corresponding to nearly 15 million frames. The dataset is collected autonomously with minimal human intervention, in a self-supervised manner, and is designed to be easily extensible to new robotic hardware, various sensors, and different collection policies. 

The common practice of re-collecting data from scratch for every new environment essentially means re-learning basic knowledge about the world --- an unnecessary effort. In this work, we show that sharing data across robots and environments makes it possible to pre-train models on a large dataset of experience, thus extracting priors that allow for fast learning with new robots and in new scenes.
If the models trained on this data can acquire the underlying shared patterns in the world, the resulting system would be capable of manipulating \emph{any} object in the dataset using \emph{any} robot in the dataset, and potentially even transfer to new robots and objects.

To learn from autonomously-collected data without explicit reward or label supervision, we require a self-supervised algorithm. To this end, we study two methods for sharing data across robot platforms and environments. First, we study the visual foresight algorithm~\cite{finn2017deep,ebert2018deep}, a deep model-based reinforcement learning method that is able to learn a breadth of vision-based robotic manipulation skills from random interaction. Visual foresight uses an
action-conditioned video prediction model trained on the collected data to plan actions that achieve user-specified goals. Second, we study deep inverse models that are trained to predict the action taken to reach one image from another image, and can be used for goal-image reaching tasks~\cite{agrawal2016learning,lynch2019play}.
However, when trained in a single environment, robot learning algorithms, including visual foresight and inverse models, do not generalize to large domain variations, such as different robot arms, grippers, viewpoints, and backgrounds, precluding the ability to share data across multiple experimental set-ups and making it difficult to share data across institutions.

Our main contributions therefore consist of the RoboNet dataset, and an experimental evaluation that studies our framework for multi-robot, multi-domain model-based reinforcement learning based on extensions of the visual foresight algorithm and prior inverse model approaches. We show that, when trained on RoboNet, we can acquire models that generalize in zero shot to novel objects, novel viewpoints, and novel table surfaces. We also show that, when these models are finetuned with small amounts of data (around 400 trajectories), they can generalize to unseen grippers and new robot platforms, and perform better than robot-specific and environment-specific training. We believe that this work takes an important step towards large-scale data-driven approaches to robotics, where data can be shared across institutions for greater levels of generalization and performance.

\section{Related Work}


Deep neural network models have been used widely in a range of robotics applications~\cite{dppt, zeng2019tossingbot,chebotar2017combining,zeng2018learning,chauffeurnet, huval2015empirical}.
However, most work in this area focuses on learning with a single robot in a single domain, while our focus is on curating a dataset that can enable a single model to generalize to multiple robots and domains.
The multi-task literature \cite{deisenroth2014multi, ammar2014online}, lifelong learning literature \cite{thrun1995lifelong, thrun1995lifelong2}, and meta-learning literature~\cite{finn2017model, alet2018modular} describe ideas that are tightly coupled with this concept. By collecting task-agnostic knowledge in wide variety of domains, a robotic system should be able to rapidly adapt to new, unseen environments using relatively little target domain data.

Large-scale, self-supervised robot learning approaches have adopted a similar viewpoint~\cite{pinto2016supersizing,levine2018learning,gupta2018robot,finn2017deep,zeng2018learning,agrawal2016learning,pathak2018zero,ebert2018deep}. Unlike these methods, we specifically consider transfer across multiple robots and environments, as a means to enable researchers to share data across institutions. We demonstrate the utility of our data by building on the visual foresight approach~\cite{finn2017deep,ebert2018deep}, as it further enables generalization across tasks without requiring reward signals. This method is related to a range of recently proposed techniques that use predictive models for vision-based control~\cite{byravan2017se3,tian2019manipulation,watter2015embed,causal_infogan, nair2018time}. Further, we also study how we can extend vision-based inverse models~\cite{agrawal2016learning,pathak2018zero,yu2019unsupervised,lynch2019play} for generalizable robot-agnostic control.

A number of works have studied learning representations and policies that transfer across domains, including transfer from simulation to the real world~\cite{sadeghi2016cad2rl,tobin2017domain,james2019sim}, transfer across different dynamics~\cite{igor,wenhao,peng2018sim,andrychowicz2018learning}, transfer across robot morphologies with invariant feature spaces~\cite{gupta2017learning} and modularity~\cite{devin2017learning}, transfer across viewpoints through recurrent control~\cite{sadeghi2018sim2real},
and transfer across
objects~\cite{finn2017one,james2018task}, tasks~\cite{duan2017one} or environments~\cite{anusha_meta} through
meta-learning. In contrast to these works, we consider transfer at a larger scale across not just one factor of variation, but across objects, viewpoints, tasks, robots, and environments, without the need to manually engineer simulated environments.


Outside of robotics, large and diverse datasets have played a pivotal role in machine learning. One of the best known datasets in modern computer vision is the ImageNet dataset~\cite{deng2009imagenet}, which popularized an idea presented earlier in the tiny image dataset \cite{torralba200880}. In particular, similar to our work, the main innovation in these datasets was not in the quality of the labels or images, but in their diversity: while prior datasets for image classification typically provided images from tens or hundreds of classes, the ImageNet Large-Scale Visual Recognition Challenge (ILSVRC) contained one thousand classes. Our work is inspired by this idea: while prior robotic manipulation methods and datasets~\cite{finn2016unsupervised,yu2016more,chebotar2016bigs,gupta2018robot,mandlekar2018roboturk,ebert2018deep,sharma2018multiple} generally consider a single robot at a time, our dataset includes 7 different robots and data from 4 different institutions,
with dozens of backgrounds and hundreds of viewpoints.
This makes it feasible to study broad generalization in robotics in a meaningful way. 

\section{Data-Driven Robotic Manipulation}
\label{sec:problem}

In this work we take a \emph{data-driven} approach to robotic manipulation. We do not assume knowledge of the robot's kinematics, the geometry of objects or their physical properties, or any other specific property of the environment. Instead, basic common sense knowledge, including rigid-body physics and the robot's kinematics, must be implicitly learned purely from data. 

\textbf{Problem statement: learning image-based manipulation skills.}
We use data-driven robotic learning for the task of object relocation -- moving objects to a specified location either via pushing or grasping and placing. 
However, in principle, our approach is applicable to other domains as well. 
Being able to perform tasks based on camera images alone provides a high degree of generality.
We learn these skills using a dataset of trajectories of images $I_{0:T}$ paired with actions $a_{0:T}$, here $T$ denotes the length of the trajectory. The actions are sampled randomly and need to provide sufficient exploration of the state space, which has been explored in prior work \cite{ebert2018deep,annie}. This learning and data collection process is self-supervised, requiring the human operator only to program the initial action distribution for data collection and to provide new objects at periodic intervals. Data collection is otherwise unattended.

\textbf{Preliminaries: robotic manipulation via prediction.}
We build on visual foresight~\cite{finn2017deep,ebert2018deep}, a method based on an action-conditioned video prediction model that is trained to predict future images, up to a horizon $h$, from on past images: $\hat{I}_{t+1:t+h} =  f(I_t, a_{t:t+h-1})$, using unlabeled trajectory data such as the data presented in the next section.
The video prediction architecture used in visual foresight is a deterministic variant of the SAVP video prediction model \cite{lee2018stochastic} based heavily on prior work~\cite{finn2016unsupervised}. This model both predicts future images and the motion of pixels, which makes it straightforward to set goals for relocating objects in the scene simply by designating points $d_{0, i}$ (e.g., pixels on objects of interest), and for each one specifying a goal position $d_{g,i}$ to which those points should be moved. We refer to $d_{0, i}$ as designated pixels. These goals can be set by a user, or a higher-level planning algorithm. The robot can select actions by optimizing over the action sequence to find one that results in the desired pixel motion, then executing the first action in this sequence, observing a new image, and replanning. This effectively implements image-based model-predictive control (MPC). With an appropriate choice of action representation, this procedure can automatically choose how to best relocate objects, whether by pushing, grasping, or even using other objects to push the object of interest. Full details can be found in Appendix~\ref{sec:visual_foresight_prelim} and in prior work~\cite{ebert2018deep}.

\textbf{Preliminaries: robotic manipulation via inverse models.}
\label{sec:inverse}
To evaluate RoboNet's usefulness for robot learning beyond use with the visual foresight algorithm, we evaluate a simplified version of the inverse model in~\cite{lynch2019play}. Given context data, $\{\dots, (I_{t-2}, a_{t-2}), (I_{t-1}, a_{t-1})\}$, the current image observation $I_t$, and a goal image $I_{t+T}$, the inverse model is trained to predict actions $a_t, \dots, a_{t+ T-1}$ (where $T$ is a given horizon) that are needed to take the robot from the start to the goal image. Our experiments train a one-step inverse model where $T=1$, which can be trained with supervised regression. At test time, the model takes as input 2 context frame/action pairs, the current image, and a goal image and then will predict an action which ought to bring the robot to the goal. This process is can be repeated at the next time-step, thus allowing us to run closed loop visual control for multiple steps.

\section{The RoboNet Dataset}
To enable robots to learn from a wide range of diverse environments and generalize to new settings, we propose RoboNet, an open dataset for sharing robot experience. An initial set of data has been collected across 7 different robots from 4 different institutions, each introducing a wide range of conditions, such as different viewpoints, objects, tables, and lighting. 
By having only loose specifications\footnote{Specifications can be found here: \url{http://www.robonet.wiki}} on how the scene can be arranged and which objects can be used, we naturally obtain a \emph{large amount of diversity}, an important feature of this dataset. By framing the data collection as a cross-institutional effort, we aim to make the diversity of the dataset grow over time. \emph{Any research lab is invited to contribute to it}. 

\subsection{Data Collection Process}
All trajectories in RoboNet share a similar action space, which consists of deltas in position and rotation to the robot end-effector, with one additional dimension of the action vector  reserved for the gripper joint. The frame of reference is the root link of the robot, which need not coincide with the camera pose. This avoids the need to calibrate the camera, but requires any model to infer the relative positioning between the camera and the robots' reference frames from a history of context frames.
As we show in Section~\ref{sec:experiments}, current models can do this effectively. The action space can also be a subset of the listed dimensions. 
We chose an action parametrization in end-effector space rather than joint-space, as it extends naturally to robot arms with different degrees of freedom.
Having a unified action space throughout the dataset makes it easier to train a single model on the entire dataset. However, even with a consistent action space, variation in objects, viewpoints, and robot platforms has a substantial effect on how the action influences the next image.

In our initial version of RoboNet, trajectories are collected by applying actions drawn at random from simple hand-engineered distributions. We most commonly use a diagonal Gaussian combined the automatic grasping primitive developed in \cite{ebert2018robustness}. More details on the data collection process are provided in Appendix~\ref{sec:app_data_coll}.

\subsection{The Diverse Composition of RoboNet}

The environments in the RoboNet dataset vary both in robot hardware, i.e. robot arms and grippers, as well as environment, i.e arena, camera-configuration and lab setting, which manifests as different backgrounds and lighting conditions (see Figure~\ref{fig:robonet_overview} and \ref{fig:attributes}). In theory, one could add any type (depth, tactile, audio, etc.) of sensor data to RoboNet, but we stick to consumer RGB video cameras for the purposes of this project. There is no constraint on the type of camera used, and in practice different labs used cameras with different exposure settings. Thus, the color temperature and brightness of the scene varies through the dataset. Object sets also vary substantially between different lab settings. To increase the number of tables, we use inserts with different textures and colors. To increase the number of gripper configurations, we 3D printed different finger attachments.
We collected 104.4k trajectories for RoboNet on a Sawyer arm, Baxter robot, low-cost WidowX arm, Kuka LBR iiwa arm, and Franka Panda arm. We additionally augment the dataset with publicly available data from prior works, including 5k trajectories from a Fetch robot~\cite{yu2019unsupervised} and 56k trajectories from a robot at Google~\cite{finn2016unsupervised}. The full dataset composition is summarized in Table~\ref{tbl:attributes}.

\subsection{Using and Contributing to RoboNet}

The RoboNet dataset allows users to easily filter for certain attributes. For example, it requires little effort to setup an experiment for training on all robots with a certain type of gripper, or all data from a Sawyer robot. An overview of the current set of attributes is shown in \autoref{tbl:attributes}, and image examples are provided in Figure~\ref{fig:attributes}. We provide code infrastructure and common usage examples on the project website.\footnote{The project webpage is at \url{http://www.robonet.wiki/} }

Scripts for controlling common types of robots, for collecting data, and for storing data in a standard format are available on the project website.
On the same webpage we are also providing a platform that allows anyone to upload trajectories. After data has been uploaded we will perform manual quality tests to ensure that the trajectories comply with the standards used in RoboNet: the robot setup should occupy enough space in the image, the action space should be correct, and the images should be of the right size.
After passing the quality test, trajectories are added to the dataset. An automated quality checking procedure is planned for future work.

\begin{minipage}[t]{\textwidth}
  \begin{minipage}[b]{0.36\textwidth}
    \centering
    \includegraphics[width=2.3\textwidth]{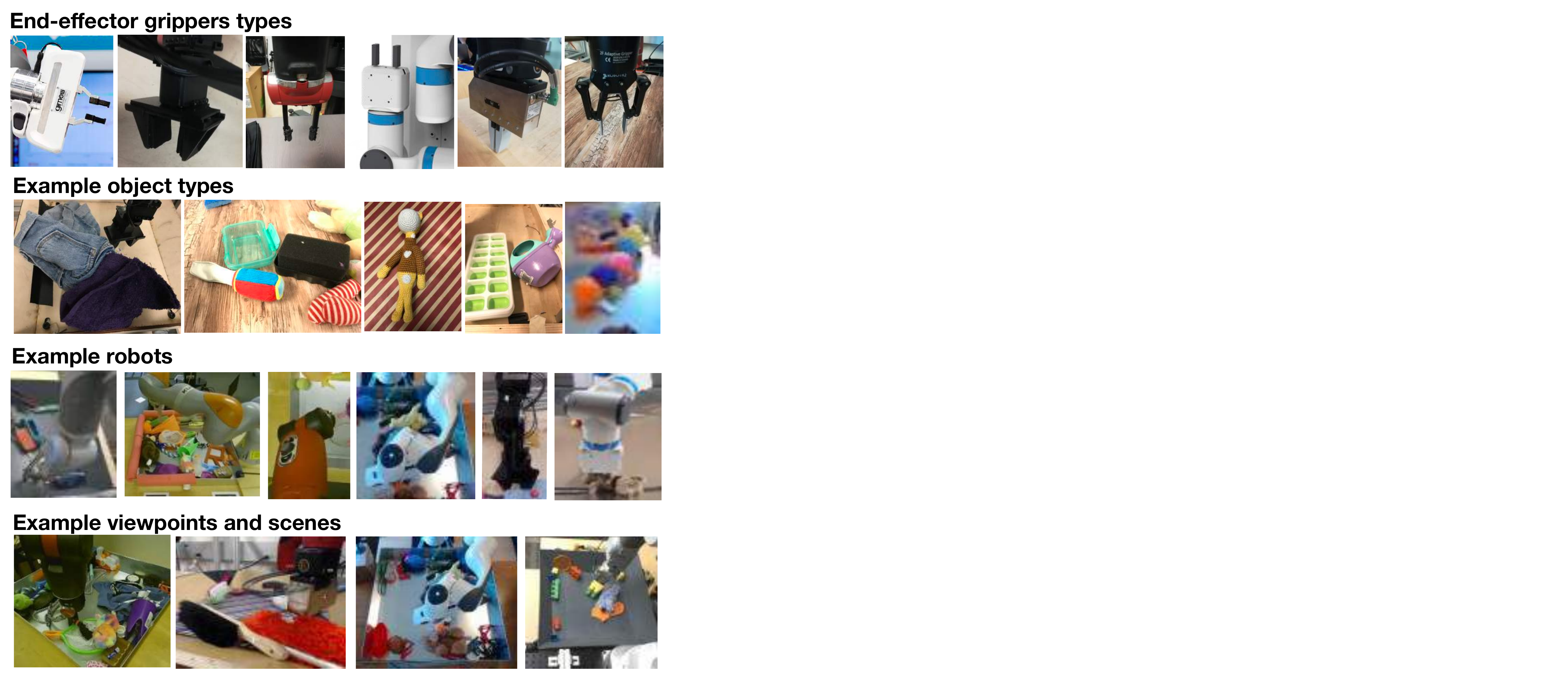}
    \captionof{figure}{Qualitative examples of the various attributes in the RoboNet dataset.
    \label{fig:attributes}}
  \end{minipage}
  \hfill
  \begin{minipage}[b]{0.62\textwidth}
    \centering
    \footnotesize
    \begin{tabular}{ ll } 
 \toprule
 \multirow{3}{4cm}{Robot type \\ {\footnotesize (number of trajectories)}} & Sawyer (68k), Baxter (18k),  \\
    &   WidowX (5k), Franka (7.9k), \\
    &  Kuka (1.8k), Fetch (5k)~\cite{yu2019unsupervised},\\ 
    & GoogleRobot (56k)~\cite{finn2016unsupervised}\!\! \\
 \hline
  \multirow{4}{3cm}{Gripper type} & Weiss Robotics WSG-50, \\
  & Robotiq, WidowX,  \\
  & Baxter, Franka, Kuka \\
 \hline
 Arena types &  7\\
 \hline
 Arena inserts & 10 \\
 \hline
 Gripper configurations & 10 \\
 \hline
 Camera configuration & 113 \\
 \hline
 Lab environments & 4 \\
 \bottomrule
\end{tabular}
      \captionof{table}{Quantitative overview of the various attributes in the RoboNet dataset, including the 6 different robot arms and 6 different grippers.
      \label{tbl:attributes}}
    \end{minipage}
  \end{minipage}




\section{Robot-Agnostic Visual Control: Model Training and Experiments}
\label{sec:experiments}


A core goal of this paper is to study the viability of large-scale data-driven robot learning as a means to acquire broad generalization, across scenes, objects, and even robotic platforms.
To this end, we design a series of experiments to study the following questions: (1) can we leverage RoboNet to enable zero-shot generalization or few-shot adaptation to novel viewpoints and novel robotic platforms?
(2) how does the breadth and quantity of data affect generalization? (3) do predictive models trained on RoboNet memorize individual contexts or learn generalizable concepts that are shared across contexts? Finally, we evaluate a simple inverse model to test if RoboNet can be used with learning algorithms other than visual foresight.

\subsection{Visual Foresight: Experimental Methodology}
For our visual foresight robot experiments, we evaluate models in terms of performance on the object relocation tasks described in Section~\ref{sec:problem}. A task is defined as moving an object not in the training set to a particular location in the image. After running the learned policy or planner, we measure the distance between the achieved object position and the goal position. We judge a task to be successful if the operator judges the object is mostly covering the goal location at the end of the rollout. Models within an experiment are compared on the same set of object relocation tasks. We use this evaluation protocol through the rest of the experiments. Please refer to Appendix \ref{sec:task_desc} for some images of the testing environments. Note that results should not be compared across different experiments, since task difficulty varies across robots and human operators.

\subsection{Visual Foresight: Zero-Shot Generalization to New Viewpoints and Backgrounds}

In this section, we study how well models trained on RoboNet can generalize, without any additional data, to novel viewpoints and held-out backgrounds with a previously seen robot. Generalizing to a new viewpoint requires the model to implicitly estimate the relative positioning and orientation between the camera and the robot, since the actions are provided in the robot's frame of reference. We attempt five different object relocation tasks from two views in order to compare a model that has been trained on 90 different viewpoints against a model that was only trained on single viewpoint. The arrangement of the cameras is shown in Appendix \ref{sec:task_desc}. In  Table~\ref{tbl:multiview}, we show object relocation accuracy results for both of these models when testing on both the seen viewpoint (left) and a novel viewpoint (right). 
The results show that the model trained on varied viewpoints achieves lower final distance to the goal on the benchmark tasks for \emph{both} views, thus illustrating the value of training on diverse datasets.

We tested the same multi-view model on a similar set of tasks in an environment substantially different from the training environment. In Figure~\ref{fig:newtable_exp} we show a successful execution of a pushing task in this new environment. The multi-view model achieves an average final distance of 14.4 $\pm$ 2 cm (std. error) in the new setting. This performance is comparable to that achieved by the multi-view model in a novel viewpoint, which suggests the model is also able to effectively generalize to novel surroundings.

\begin{minipage}[t]{\textwidth}
  \begin{minipage}[b]{0.5\textwidth}
	\centering
            	\includegraphics[width=1.0\columnwidth]{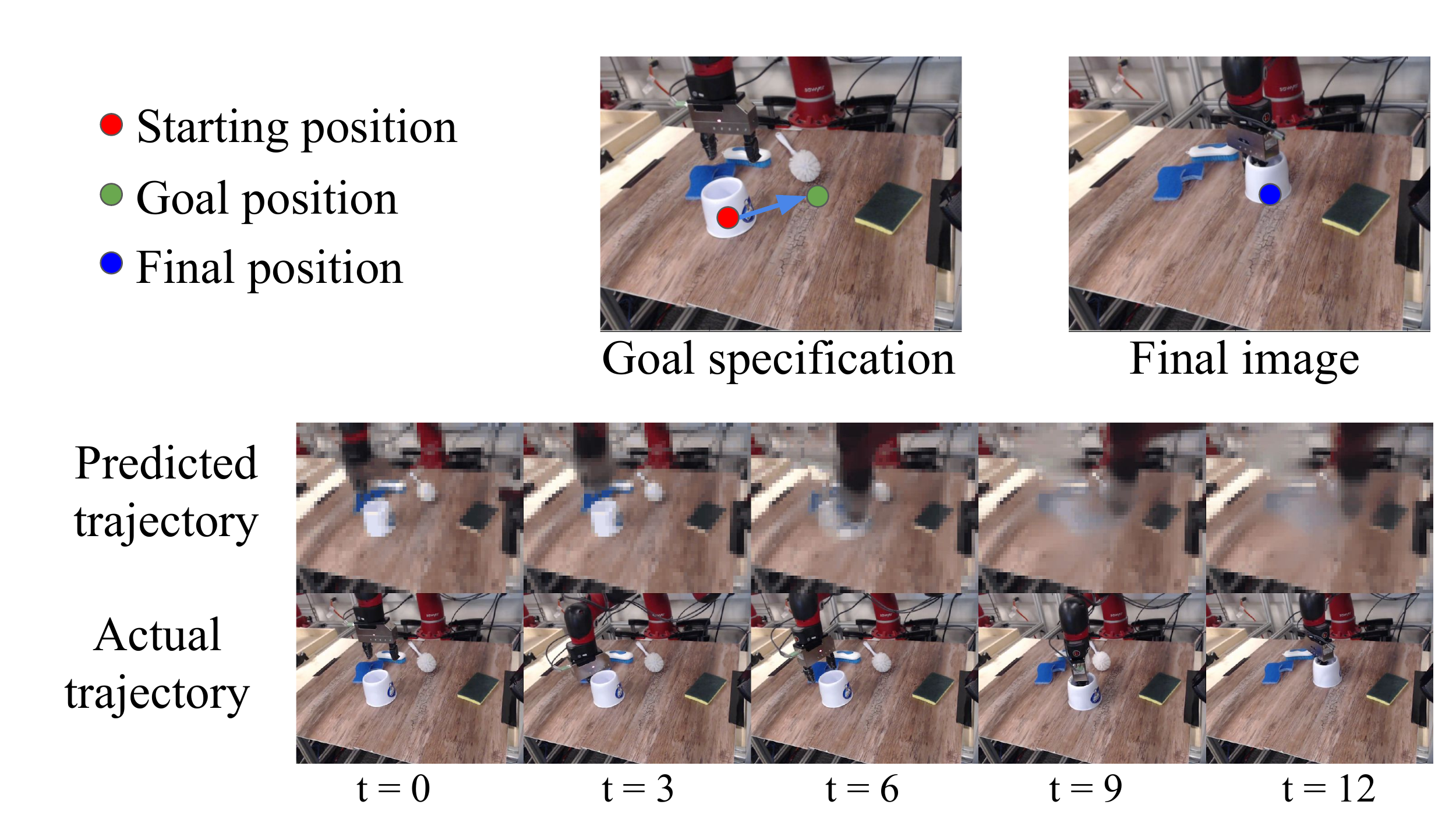}
	\vspace{-0.4cm}
	\captionof{figure}{\small{Zero-shot generalization to new backgrounds with a model trained across multiple views. }
	\label{fig:newtable_exp}}
  \end{minipage}
  \hfill
  \begin{minipage}[b]{0.45\textwidth}
  \vspace{-0.55in}
    \centering
    \footnotesize
\vspace{-0.1in}
\begin{tabular}{ lcc } 
 \toprule
  & \!\!\!\!\thead{Avg. dist. (cm) \\ seen view} & \!\!\!\!\thead{Avg. dist.  (cm) \\ held-out view} \\
 \hline
 single view & 14.8 $\pm$	3.8 & 23.2	$\pm$ 2.6  \\
 \hline
 multi-view  & \textbf{9 $\pm$	2.2 } &  \textbf{16.2$\pm$ 	2.9}\\
 \bottomrule
\end{tabular}
 \captionof{table}{\footnotesize Evaluation of viewpoint generalization, showing the average distance to the goal after executing the action sequence and standard error. A model trained on multiple views can better generalize to a new viewpoint.}
\label{tbl:multiview}
\end{minipage}
\end{minipage}

\subsection{Visual Foresight: Few-Shot Adaptation to New Robots}

When evaluating on domains that differ more substantially from any domain present in the dataset, such as settings that contain an entirely new robotic arm, zero-shot generalization is not possible. In this section, we evaluate how well visual foresight can adapt to entirely new robots that were not shown to the model during training. This is one of the most challenging forms of generalization, since robots have not only different appearances, but also different dynamics when interacting with objects, different kinematics, and different work-space boundaries.

To test our hypothesis, we collect a small number (300-400) of random trajectories from the target robot environment. Models are then pre-trained on the entirety of RoboNet, but holding out the data from the target robot. These models are then fine-tuned using the aforementioned collected trajectories. We compare to a separate model that is trained from scratch on those trajectories. Additionally, for the Franka experiments another model is trained on all the Franka data in RoboNet, and for the Baxter experiment one model is pre-trained on just Sawyer data in RoboNet and fine-tuned to Baxter. The R3 and Fetch were also not included in the pre-training data due to computational constraints.

\begin{minipage}[t]{\textwidth}
  \begin{minipage}[b]{0.32\textwidth}
    \centering
    \footnotesize

\begin{tabular}{ lcc } 
 \toprule \textbf{Kuka Experiments}
   & \thead{Success\!\!\\ rate}  \\
 \hline
 \textit{Random Initialization} \\ Train on N=400 \!\!  & 10\% \\
 \hline
\textit{Random Initialization} \\ Train on N=1800   &  30\% \\
\hline
\textit{Pre-train on RoboNet} \\ \textit{w/o Kuka, R3, Fetch} \\ Finetune on N=400 & \textbf{40\%} \\
 \bottomrule
\end{tabular}
 \captionof{table}{\footnotesize{Results for adaptation to an unseen Kuka robot. The model pre-trained on RoboNet without the Kuka, R3, and Fetch data, achieves the best performance when fine-tuned with 400 trajectories from the test robot.}
\label{tbl:finetune_kuka}}
\end{minipage}
  \hfill
\begin{minipage}[b]{0.32\textwidth}
    \centering
    \footnotesize
\begin{tabular}{ lcc } 
 \toprule \textbf{Franka Experiments}
   & \thead{Success\!\!\\ rate}  \\
 \hline
 \textit{Random Initialization} \\ Train on N=400 \!\!  & 20\% \\
 \hline
\textit{Random Initialization} \\ Train on N=8000   &  35\% \\
\hline
\textit{Pre-train on RoboNet} \\ \textit{w/o Franka, R3, Fetch} \\ Finetune on N=400 & \textbf{40\%} \\
 \bottomrule
\end{tabular}
 \captionof{table}{\footnotesize{Results for adaptation to an unseen Franka robot. The model pre-trained on RoboNet without the Franka, R3, and Fetch data, achieves the best performance when fine-tuned with 400 trajectories from the test robot.}
\label{tbl:finetune_franka}}
\end{minipage}
  \hfill
  \begin{minipage}[b]{0.32\textwidth}
    \centering
    \footnotesize
\begin{tabular}{ lcc } 
 \toprule \textbf{Baxter Experiments}
  & \thead{Success\!\!\\ rate}  \\
 \hline
 \textit{Random Initialization} \\ Train on N=300  & 33\% \\
 \hline
\textit{Pre-train on Sawyer} \\ Finetune on N=300 & \textbf{83\%} \\
\hline
\textit{Pre-train on RoboNet} \\ \textit{w/o Baxter} \\ Finetune on N=300 & 58\% \\
 \bottomrule
\end{tabular}
 \captionof{table}{\footnotesize{Evaluation results for adaptation to an unseen Baxter robot. The model pre-trained on RoboNet's Sawyer data, achieves the best performance when fine-tuned with 300 trajectories from the test robot.}
\label{tbl:finetune_baxter}}
\end{minipage}
\end{minipage}



\begin{wrapfigure}{r}{.5\columnwidth}
	\centering
	\includegraphics[width=0.5\columnwidth]{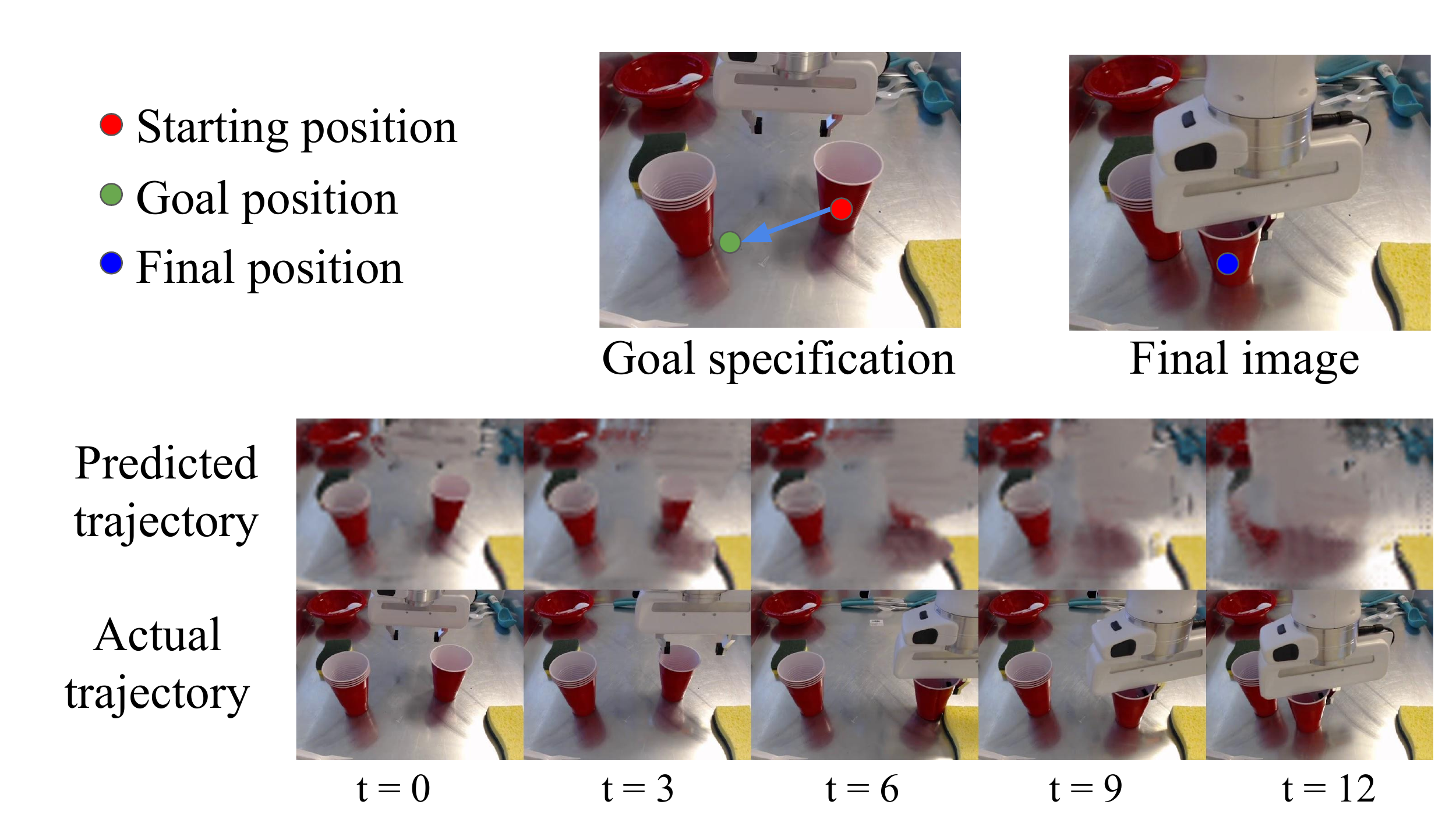}
	\vspace{-0.5cm}
	\captionof{figure}{\small{Example task of grasping and moving a thin plastic cup with the Franka robot, using visual foresight pre-trained on RoboNet w/o Franka and fine-tuned on 400 trajectories from the Franka robot.}
	\label{fig:fintune_franka}}
	\vspace{-0.4cm}
 \end{wrapfigure}
The quantitative results are summarized in  \autoref{tbl:finetune_kuka}, \autoref{tbl:finetune_franka}, and \autoref{tbl:finetune_baxter}. 
The results show that RoboNet pre-training provides substantial improvements over training from scratch, on all three test robots. In the Kuka and Franka experiments, a model fine-tuned on \textit{just 400 samples} is able to outperform its counterpart trained on all of RoboNet's data from the respective robot. These results suggest that RoboNet pre-training can offer large advantages over training tabula rasa, by substantially reducing the number of samples needed in a new environment.  \autoref{fig:fintune_franka} shows a successful rollout of visual foresight on a challenging task of positioning a plastic cup to a desired location.

In the Baxter experiment, we also find that pre-training on specific subsets of RoboNet (in this case the Sawyer, which is visually more similar to the Baxter than other robots) can perform significantly better than training on the entire dataset.
Hence, this experiment (as well as the Robotiq gripper generalization experiment in Appendix \ref{sec:robotiq}) demonstrates that increased diversity during pre-training can sometimes hurt performance when compared to pre-training on a subset of RoboNet. We hypothesize that more specific pre-training works better, because our models under-fit when trained on all of RoboNet, which we study in more detail in the next section.

\subsection{Visual Foresight: Model Capacity Experiments}
\label{sec:underfitting}



When training video prediction models on RoboNet, we observe clear signs of under-fitting. Training error and validation error are generally similar, and both plateau before reaching very high performance on the training sequences. During test time, inaccurate predictions are often the cause of poor performance on the robot. Thus, we perform an additional experiment to further validate the underfitting hypothesis. We train two large models, using a simplified deterministic version of the network architecture presented in~\cite{TPU}, on RoboNet's Sawyer data: one model has 200M parameters and the other has 500M parameters. The 200M parameter model has $0.104 \pm 0.057$ average $\ell_1$ per-pixel error on a held out test set, whereas the 500M model has $0.0847 \pm 0.045$ $\ell_1$ per-pixel error. These results suggest that current visual foresight models -- even ones much larger than the 5M - 75M parameter models used in our control experiments -- suffer from underfitting, and future research on higher capacity models will likely improve performance.


\subsection{Inverse Model: Multi-Robot and Multi-Viewpoint Reaching}
\label{sec:Inverse Model}

\begin{wrapfigure}{r}{.34\columnwidth}
\vspace{-0.2cm}
    \centering
    \footnotesize
\begin{tabular}{ lc } 
 \toprule \thead[l]{\textbf{Inverse Model}}
   & \thead{Success\!\!}  \\
 \hline
\textit{Sawyer Reaching} \\ Front View  &  4/5 \\
\hline
\textit{Sawyer Reaching} \\ Unseen View &  5/5 \\
\hline
\textit{Franka Reaching} \\ Front View & 4/5 \\

 \bottomrule
\end{tabular}
 \captionof{table}{\footnotesize{Inverse model results on 5 reaching tasks. The model is successful across multiple robot platforms and generalizes to a new viewpoint.}
 \vspace{-0.4cm}
\label{tbl:inverse_table}}
\end{wrapfigure}

To evaluate RoboNet's applicability to different control algorithms, we train a simple version of the inverse model from \cite{lynch2019play}
(refer to Section~\ref{sec:inverse} for details) on a subset of RoboNet containing only Sawyer and Franka data. The \textit{same} model is evaluated on both robots: the Sawyer experiments also contain a held-out view. We evaluate model performance on simple reaching tasks. Tasks are constructed by supplying a goal image, by taking an image of the gripper in a different reachable state. After task specification, the model runs continuously, re-planning each step until a maximum number of steps is reached. Success is determined by a human judge. This model is able to perform visual reaching tasks on both robots, including from a novel viewpoint not seen during training. However, because of its comparatively greedy action selection procedure, we observe that it tends to perform poorly on more complex tasks that require object manipulation.


\section{Discussion}
\label{sec:conclusion}

We presented RoboNet, a large-scale and extensible database of robotic interaction experience that combines data from 7 different robots, multiple environments and backgrounds, over a hundred camera viewpoints, and four separate geographic locations. We demonstrated two example use-cases of the dataset by (1) applying the visual foresight algorithm~\cite{ebert2018deep} and (2) learning vision-based inverse models. We evaluated generalization across many different experimental conditions, including varying viewpoints, grippers, and robots. Our experiments suggested that fine-tuning models pretrained on RoboNet offers a powerful way to quickly allow robot learning algorithms to acquire vision-based skills on unseen robot hardware.

Our experiments further found that video prediction models with $\leq$ 75M parameters tend to heavily \emph{underfit} on RoboNet. While much better, we even observe underfitting on 500M-parameter models. As a result, prediction models struggle to take advantage of the breadth and diversity of data from multiple robots, domains, and scenes, and instead seem to perform best when using a subset of RoboNet that looks most similar to the test domain.
This suggests two divergent avenues for future work. On one hand, we can develop algorithms that automatically select subsets of the dataset based on various attributes in a way that maximizes performance on the test domain. In the short term, this could provide considerable improvements with our current models. 
However, an alternative view is to instead research how to build more flexible models and policies, that are capable of learning from and larger and more diverse datasets across many robots and environments. We hope that the RoboNet dataset can serve as a catalyst for such research, enabling robotics researchers to study such problems in large-scale learning.
Next, we discuss limitations of the dataset and evaluation, and additional directions for future work.


\textbf{Limitations.} While our results demonstrated a large degree of generalization, a number of important limitations remain, which we aim to study in future work. First and foremost, the tasks we consider are relatively simple manipulation tasks such as pushing and pick-and-place, with relatively low fidelity. This is an important limitation that hinders the ability of these models to be immediately of practical use.
However, there are a number of promising recent works that have demonstrated how predictive models of observations can be used for solving tasks of greater complexity such as tool use~\cite{annie} and rope manipulation~\cite{causal_infogan}, and tasks at greater fidelity such as block mating~\cite{nair2018time} and die rolling~\cite{tian2019manipulation}. Further, one bottleneck that likely prevents better performance is the quality of the video predictions. We expect larger, state-of-the-art models ~\cite{video_transformer, TPU} to produce significantly better predictions, which would hopefully translate to better control performance.

Another limitation of our current approach and dataset is the source of data being from a pre-determined random policy. This makes data collection scalable, but at the cost of limiting more complex and nuanced interactions. In future work, we plan to collect and solicit data from more sophisticated policies. This includes demonstration data, data from modern exploration methods that scale to pixel observations~\cite{bellemare2016unifying,rnd,pathak2019self}, and task-driven data from running reinforcement learning on particular tasks. As shown in prior work~\cite{annie}, improving the forms of interactions in the dataset can significantly improve performance. 

In selecting how and where to collect additional data, our experiments suggest that adaptation to new domains is possible with only modest amounts of data, on the order of a few hundred trajectories. This suggests that prioritizing variety, i.e. small amounts of data from many different domains, is more important than quantity in future collection efforts.


\textbf{Future Directions.} This work takes the first step towards creating learned robotic agents that can operate in a wide range of environments and across different hardware. While in this work, we explored two particular classes of approaches, we hope that RoboNet will inspire the broader robotics and reinforcement learning communities to investigate how to scale model-based \emph{or} model-free RL algorithms to meet the complexity of the real world, and to contribute the data generated from their experiments back into a shared community pool. In the long term, we believe this process will iteratively strengthen the dataset, and thus allow the algorithms derived from it to achieve greater levels of generalization across tasks, environments, robots, and experimental set-ups. 



\acknowledgments{
We thank Dr. Kostas Daniilidis for contributing lab resources and funding to the project. We also thank Annie Xie for discussion and help with implementation in the early stages of this project. This research was supported in part by the National Science Foundation under IIS-1651843, IIS-1700697, and IIS-1700696, the Office of Naval Research, ARL DCIST CRA W911NF-17-2-0181, DARPA, Berkeley DeepDrive, Google, Amazon, and NVIDIA.
}


\newpage
\begin{footnotesize}
\bibliography{references}  
\end{footnotesize}

\newpage
\appendix

{\footnotesize
\section{Visual Foresight Preliminaries}
\label{sec:visual_foresight_prelim}

Here we give a brief introduction into the visual foresight algorithm used in this paper, see \cite{finn2017deep, ebert2017self, ebert2018robustness} for a more detailed treatment. 

\subsection{Action conditioned video-prediction model}

The core of visual foresight is the action conditioned video-prediction model, which is a deterministic variant of the model described in \cite{lee2018stochastic}. The model is illustrated in \autoref{fig:vidpred_model} and implemented as a recurrent neural network using actions $a_{0:T}$, and images $I_{0:T}$ as inputs and outputting future predicted images $\hat{I}_{1:T}$. Instead of using a context of 1 as shown in \autoref{fig:vidpred_model}, a longer context can be used which increases the model's ability to adapt to environment variation such as held-out view-points. In all experiments in this paper we used a context of 2 frames. Longer contexts can potentially help the model adapt to unseen conditions in the environment, however, this is left for future work.
The RNN is unrolled according to the following equations:

\begin{align}
[h_{t+1}, \hat{F}_{t+1 \leftarrow t}] 	&= g_{\theta}(a_t, h_t, I_t) \\
\hat{I}_{t+1} 							&= \hat{F}_{t+1 \leftarrow t} \diamond  \hat{I}_t 
\label{simple_dna}
\end{align}
Here $g_{\theta}(a_t, h_t, I_t)$ is a forward predictor parameterized by $\theta$ and $\hat{F}_{t+1 \leftarrow t}$ is two-dimensional flow field with the same size as the image which is used to transform an image from one time-step into the next via bilinear-sampling.

The architecture of the RNN, which is illustrated in \autoref{fig:vidpred_arch}, uses a stack of convolutional LSTMs~\cite{xingjian2015convolutional} interleaved with convolution layers, skip connection help the learning process.

\begin{figure}[b]
    \centering
    \includegraphics[width=0.6\textwidth]{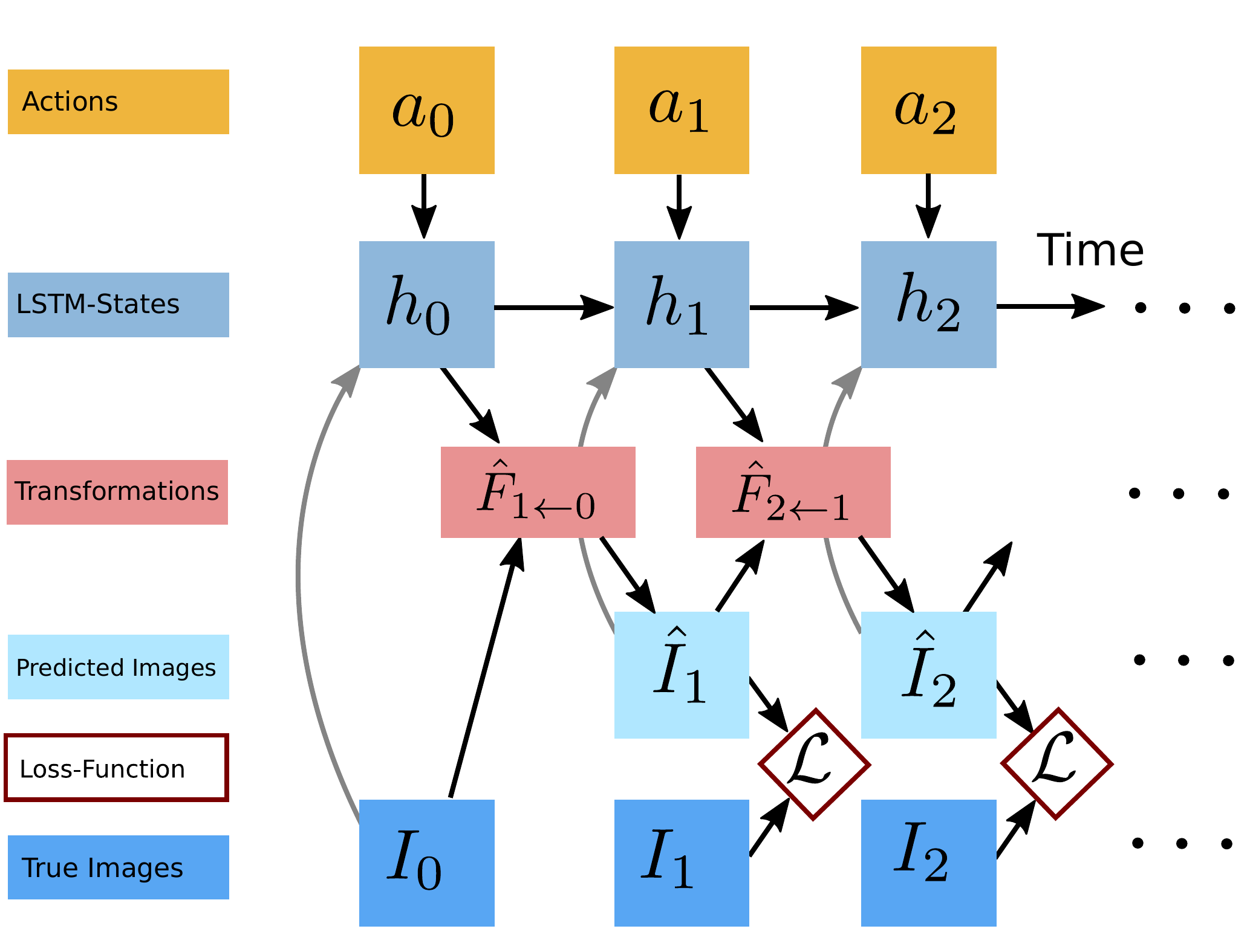}
    \caption{\small{Recurrent dynamics model for action-conditioned video-prediction based on flow transformations. (Used with permission from \cite{ebert2018deep})}}
    \label{fig:vidpred_model}
\end{figure}

\begin{figure}
    \centering
    \includegraphics[width=0.8\textwidth]{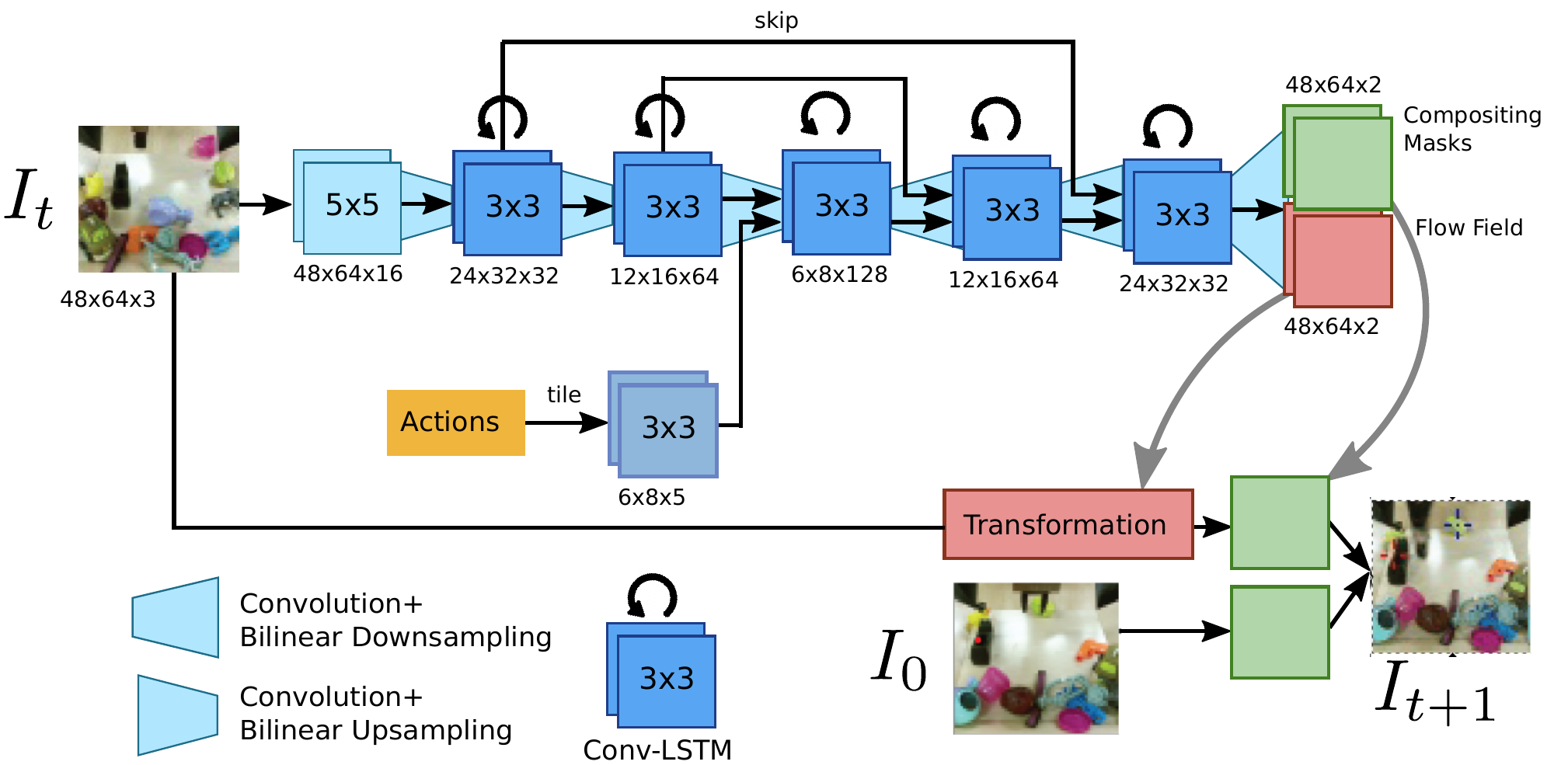}
    \caption{\small{Architecture of the recurrent video-prediction architecture. (Used with permission from \cite{ebert2018deep})}}
    \label{fig:vidpred_arch}
\end{figure}

\textbf{Training details}
For pretraining all models are trained for 160k iterations using a batchsize of 16. For SGD we use the Adam optimizer. Finetuning is carried out for another 150k steps. The learning rate starts at 1e-3 and is annealed linearly to 0 after 200k steps until the end of training.

\subsection{Sampling-based Planning}

In visual foresight tasks are specified in terms of the motion of user-selected pixels. To predict where pixels move in response to a sequence of actions, we define a probability distribution $P_0$ over the locations of the designated pixel. At time step 0 this we use a one-hot-distribution with 1 a the user-selected pixel and 0 everywhere else. When then apply the same transformations to these distributions that we also apply to the images. 
This is summarized in the following equation:
\begin{equation}
\hat{P}_{t+1} = \hat{F}_{t+1 \leftarrow t} \diamond  \hat{P}_t
\label{eqn:prob_forward}
\end{equation}
Here $\hat{P}_{t+1}$ denotes the predicted probability distribution of the designated pixel.

The planning cost is computed as the expected distance to the goal pixel position $d_g$ under the predicted distribution $\hat{P}_t$, averaged over time:
 \begin{align}
c = \sum_{t = 1, \dots, T} c_t =  \sum_{t = 1, \dots, T} \mathbb{E}_{\hat{d}_{t} \sim P_{t}} \left[\|\hat{d}_{t} - d_{g}\|_2\right] 
 \label{eq:cost}
 \end{align}
 
To find the optimal action sequence we apply the model-predictive path intregral (MPPI)~\cite{mppi} algorithm, since this allows us to find good actions sequences more efficiently than random shooting. In the first iteration the actions are sampled from a unit Gaussian, in subsequent iterations the mean action is computed as an exponential weighted average as follows:
\begin{equation}
    \mu_t = \frac{\sum^N_{k=0} e^{-\gamma c_k} a_{k, 0:T}}{\sum^N_{k=0} e^{-\gamma c_k}}
\end{equation}
Here $N$ is the number of samples, chosen to be 600. The prediction horizon is 15 steps. We found that a number of 3 MPPI iterations works best in our settings. We apply temporal smoothing to the action samples using a low-pass filter to achieve smoother control and better exploration of the state space.

After finding an action sequence, the first action of this sequence is applied to the robot and the planner is queried again, thus operating in an MPC-like fashion. In order to perform re-planning, it is required to know the current position of the designated pixel. In this work we use a simple method for obtaining an estimate for the designated pixel by using the model's prediction, i.e. the flow maps from the previous time-step, we call this predictor propagation. While this position estimate is noisy and more complex alternatives, such as hand-engineered trackers or self-supervised registration \cite{ebert2018robustness} exist, we opt for the simple approach in this work.

\section{Data Collection Details}
\label{sec:app_data_coll}
\subsection{State and Action Space}
Most of the robots in RoboNet (excluding Google R3 from \cite{finn2016unsupervised}) employ the same Cartesian end-effector control action space with restricted rotation, and a gripper joint. At each time-step, the state is a $\mathbb{R}^5$ vector containing the grippers XYZ position, the gripper's yaw angle (rest of orientation is locked, with the gripper pointed downwards), and the gripper joint-angle value. The user must specify safety bounds per-robot, which constrain the end-effector to operate within a "safe" region of space at all time-steps. Actions are specified as deltas between the current state and commanded state for the next time-step. Note that the gripper action is binarized to "open" or "close" for simplicity. Actions are blocking with a set time-out, so user defined policies only receive states and calculate actions once the robot  has reached (or gotten as close as possible to) the commanded state. There are no "real-time" constraints on the user policy. As a result, the robot must come to a complete stop at each step. While this scheme can easily generalize to new robots, it does impose constraints on the final robot behavior. We hope to relax these constraints in future work.

\subsection{Exploration Policy}
During data collection, actions are either sampled from a simple diagonal Gaussian with one dimension per action-space dimension, or a more sophisticated distribution that biases the probability of grasping when the gripper is at the table height, increasing the chance that the robot will randomly grasp objects. This primitive is described further below. The variances in the diagonal Gaussians are hand-tuned per robot and differ between different action dimensions. The exact parameters are stored in inside the hdf5-files under the field \texttt{policy-description}.

While using a simple action distribution such as a diagonal Gaussian, the robot arm frequently pushes objects, however the arm quite rarely grasps objects. In order for a learning algorithm such as visual foresight to effectively model grasping, it must have seen a sufficient number of grasps in the dataset. By applying a grasping primitive, such as the one originally introduced in \cite{ebert2018robustness}, the likelihood for these kinds of events can be increased. The grasping primitive is implemented as a hard-coded rule that closes the gripper when the $z$-level of the end-effector is less than a certain threshold, and opens the gripper if the arm is lifted above a threshold while not carrying an object.

There are, however, two robots in this dataset which employ significantly different exploration policies. The Google R3 robot samples random pushing motions instead of simply taking random Cartesian motions, and the Fetch robot data only contains random exploration in the $x$ and $y$ dimensions.

\section{Database Implementation Details}

The database stores every trajectory as a separate entity with a set of attributes that can be filtered. We provide code infrastructure that allows a user to filter certain subsets of attributes for training and testing.
The database can be accessed using the Pandas python-API, a popular library for structuring large amounts of data. Data is stored in the widely adopted \texttt{hdf5}-format, and videos are encoded via MP4 for efficiency reasons. New trajectory attributes can be added easily. 

\section{Description of Benchmarking Tasks}
\label{sec:task_desc}
For all control benchmarks we used object relocation tasks from a set of fixed initial positions towards a set of fixed goal positions marked on a table. The experimental setups for each robot are depicted in \autoref{fig:physical_setups}. 
After executing the action sequences computed by the algorithm the remaining distance to the goal is measured using a tape, and success is determined by human judges.

	\includegraphics[width=1.0\columnwidth]{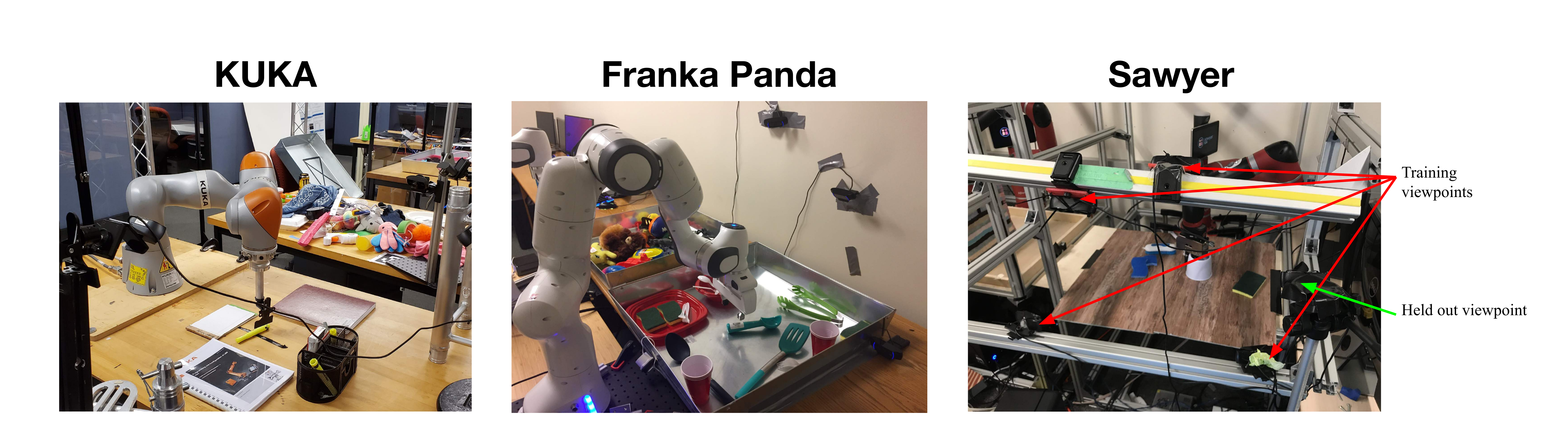}
	\vspace{-0.4cm}
	\captionof{figure}{\small{Experimental setups for benchmarking tasks on the Kuka, Franka, and Sawyer robots. }
	\vspace{0.4cm}
	\label{fig:physical_setups}}

\section{Experimental evaluation of adaptation to unseen gripper}
\label{sec:robotiq}
\vspace{0.2cm}

We evaluate on a domain where a Sawyer robot is equipped with a new gripper that was not seen in the dataset. We collected 300 new trajectories with a Robotiq 2-finger gripper, which differs significantly in visual appearance and dimensions from the Weiss Robotics gripper used in all other Sawyer trajectories (see \autoref{fig:attributes}), and used this data to evaluate four different models: zero-shot generalization for a model trained on RoboNet, a model trained only on the new data, a model pre-trained on only the Sawyer data in RoboNet and then finetuned with the new data, and a model pre-trained on all of RoboNet and finetuned with the new data. The results qualitative results of this evaluation are shown in \autoref{fig:robotiq_qual} and the quantitative results are in \autoref{tbl:adapt_to_gripper}, averaging over 10 trajectories each. The best-performing model in this case is the one that is pretrained on only the Sawyer data, and it attains performance that is comparable to in-domain generalization (see, e.g., the seen viewpoint result in \autoref{tbl:multiview} for a comparison). The model pre-trained on the more diverse RoboNet dataset actually performs worse, likely due to the limited capacity and underfitting issues discussed in Section~\ref{sec:underfitting}. However, these results do demonstrate that visual foresight models can adapt to moderate morphological changes using a modest amount of data.


\begin{minipage}[t]{\textwidth}
  \begin{minipage}[b]{0.5\textwidth}
	\centering
	\includegraphics[width=1.0\columnwidth]{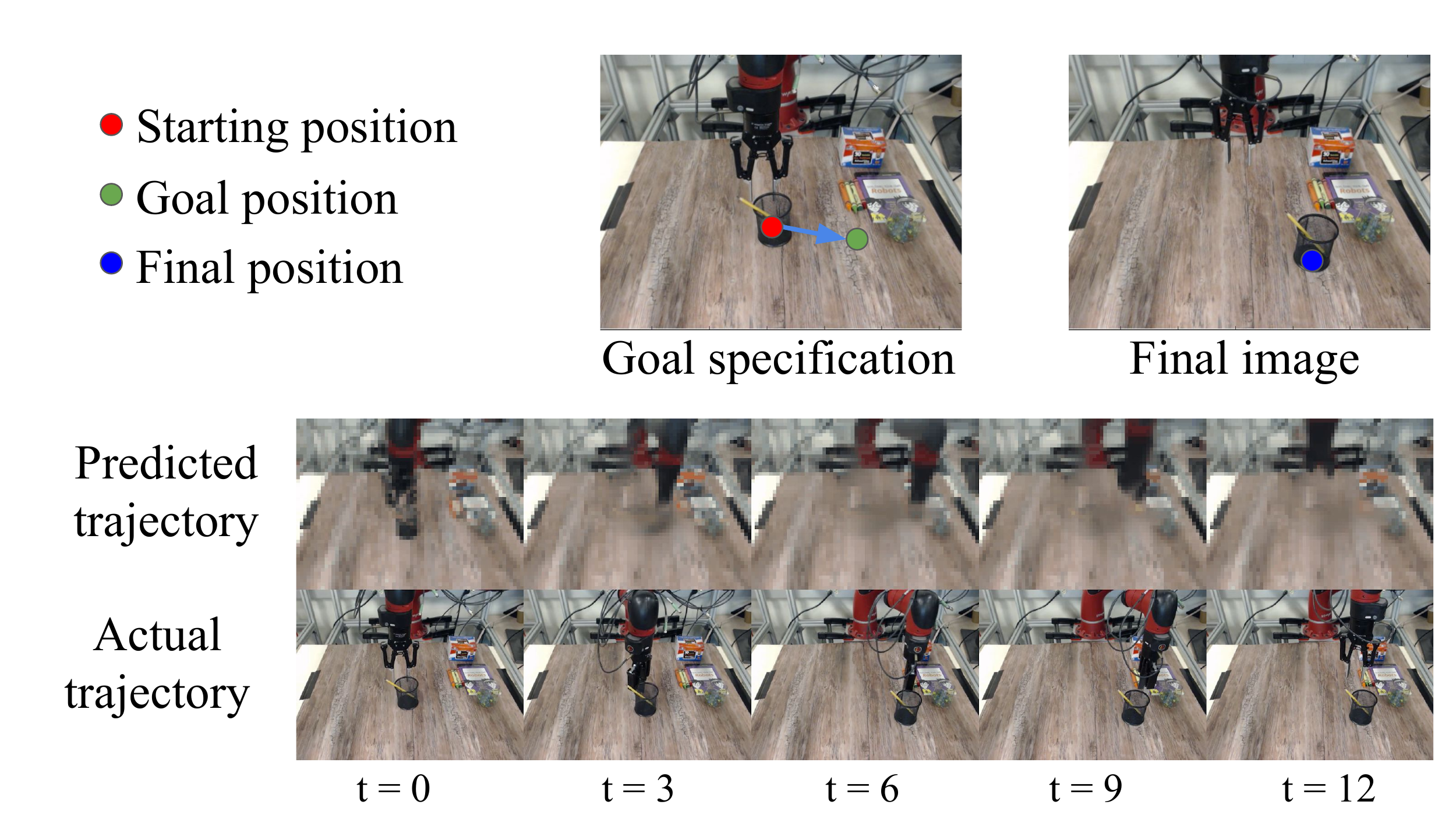}
	\vspace{-0.4cm}
	\captionof{figure}{\small{Example task of pushing an object with an unseen gripper, in this case the Robotiq gripper.}
	\label{fig:robotiq_qual}}
  \end{minipage}
  \hfill
  \begin{minipage}[b]{0.45\textwidth}
  \vspace{-0.55in}
    \centering
    \footnotesize
\vspace{-0.1in}
\begin{tabular}{ lc } 
 \toprule
  & Avg. distance (cm)    \\
   \hline
 zero-shot & 15.5 $\pm$	2.6\\
 \hline
 without pretraining & 17 $\pm$	1.8 \\
 \hline
pretraining on \\ Sawyer-only  & 	\textbf{9.8	 $\pm$ 2.1} \\
\hline
pretraining on \\ all of RoboNet  & 14.7 $\pm$	2.1 \\
 \bottomrule
\end{tabular}
 \captionof{table}{\footnotesize Evaluation results for adaptation to Robotiq gripper with the Sawyer arm. The model trained on only Sawyer data performs the best when fine-tuned on 300 trajectories with a Robotiq gripper.}
\label{tbl:adapt_to_gripper}
\end{minipage}
\end{minipage}
}

\end{document}